\documentclass[runningheads]{llncs}
\usepackage[T1]{fontenc}
\usepackage{graphicx}
\usepackage{array}
\usepackage{makecell} 
\usepackage{rotating}
\usepackage{amsmath}
\usepackage{amssymb}
\usepackage{tabularx}
\usepackage{makecell}
\usepackage{multirow}
\usepackage{adjustbox}
\usepackage{microtype}

\usepackage[colorlinks,urlcolor=black,linkcolor=blue,citecolor=blue]{hyperref}

\usepackage{xcolor}
\usepackage[table]{xcolor}
\definecolor{darkgreen}{RGB}{0,153,0}
\definecolor{darkred}{RGB}{192,0,0}
\definecolor{green1}{RGB}{222,255,226}
\definecolor{orange1}{RGB}{255,207,156}
\definecolor{red1}{RGB}{240,180,180}
\begin{document}
\title{From Visual Attribution to Clinical Reasoning:
Explainable Parkinson’s Disease Screening from
Hand-Drawn Patterns}
\titlerunning{From Visual Attribution to Clinical Reasoning ...}
%

\author{Aritra Dey\inst{1}\orcidID{0000-0001-6533-1183} \and
Utsav Kumar Nareti\inst{2}\orcidID{0000-0002-1578-371X} \and 
Chandranath Adak\inst{2}\orcidID{0000-0002-9085-2770} \and
Soumi Chattopadhyay \inst{1}\orcidID{0000-0002-9231-4087} \and
Krishna Gopal Sasmal \inst{3} \and
Saeed Anwar \inst{4} \orcidID{0000-0002-0692-8411}
} 
\authorrunning{A. Dey et al.}
\institute{
Dept. of CSE, Indian Institute of Technology Indore, India-453552 \and
Dept. of CSE, Indian Institute of Technology Patna, India-801106 \and
M. R. Bangur Hospital, West Bengal, India-700033 \and 
University of Western Australia, Australia-6009\\
\email{aritradey1993@gmail.com, chandranath@iitp.ac.in, soumi@iiti.ac.in}
}

\maketitle              


\begin{abstract}
Parkinson’s disease (PD) manifests early neuromotor impairments that become observable in controlled hand-drawn patterns such as spirals and meanders, where tremor-induced oscillations, stroke irregularity, and curvature instability reflect underlying motor degradation. In this work, we present an explainable framework for PD screening from offline hand-drawn patterns that integrates discriminative visual modeling with clinically grounded reasoning. The predictive model captures distributed structural distortions and fine-grained texture variations. It is evaluated under subject-disjoint protocols to ensure reliable generalization. To move beyond black-box classification, we introduce a multi-stage explainability pipeline that combines visual attribution with structured symptom abstraction. Salient regions are identified using attention- and gradient-based localization, followed by extraction of clinically meaningful motor descriptors quantifying contour roughness, curvature irregularity, stroke variability, and tremor-frequency energy. These descriptors are subsequently translated into coherent clinical rationales through a language-based reasoning module, linking model evidence to established PD symptomatology. By bridging visual attribution and clinical interpretation, the proposed framework advances interpretable document intelligence for neurological screening using hand-drawn patterns. Experimental results on publicly available Parkinson’s disease handwriting datasets demonstrate competitive predictive performance and clinically consistent explanations.

\end{abstract}

\section{Introduction}

Parkinson’s disease (PD) is a progressive neurodegenerative disorder characterized by bradykinesia, rigidity, tremor, and postural instability. First described by James Parkinson in 1817 \cite{doi:10.1176/jnp.14.2.223}, PD remains challenging to diagnose at early stages, where subtle neuromotor impairments often precede definitive clinical confirmation. Motor abnormalities associated with early-stage PD are well documented \cite{TOLOSA2021385}, and symptoms such as bradykinesia, rigidity, and tremor directly affect fine motor control during drawing tasks. Consequently, abnormalities in hand-drawn spirals and meanders, including tremor-induced oscillations, stroke irregularity, and curvature instability, have emerged as promising indicators of impaired motor planning and execution. Clinical studies \cite{https://doi.org/10.1002/mds.29796,https://doi.org/10.1002/mdc3.14174} suggest that although early diagnosis cannot reverse disease progression, it enables timely therapeutic intervention and improved disease management. Furthermore, the global burden of PD continues to rise, with approximately 11.77 million individuals living with the disease in 2021 compared to 6.1 million in 2016 \cite{Lie095610}. Limited access to neurological expertise, particularly in rural and resource-constrained regions, often results in delayed diagnosis and underreporting. Therefore, automated analysis of simple hand-drawn patterns offers a scalable, accessible, and cost-effective approach for early PD risk assessment and clinical decision support.

From a computational perspective, traditional PD detection methods rely on handcrafted geometric and frequency-domain features extracted from spiral and meander drawings, followed by classical machine learning classifiers. More recent approaches employ CNNs \cite{gazda2022multiple,KAMRAN2021234} and transformer-based architectures \cite{liang2024transfer} to learn visual representations directly from image data. While these methods achieve promising predictive performance, their clinical interpretability remains limited. Most existing approaches explain predictions using saliency maps, attention visualization, or feature importance scores, which indicate where the model focuses but provide little insight into the underlying motor abnormalities associated with Parkinsonian handwriting. 
In clinical settings, interpretability is essential. Standard explanation techniques such as saliency maps, Grad-CAM, and attention visualization provide localized importance cues but fail to translate visual evidence into clinically meaningful reasoning. For PD screening, clinicians require explanations expressed in terms of measurable motor characteristics, such as clinically relevant motor descriptors, rather than isolated pixel-level heatmaps. Bridging this gap between visual attribution and clinical reasoning remains an open challenge in document-based medical AI. 

To address this limitation, we propose an explainable framework that systematically connects visual attribution to structured clinical interpretation. The framework first identifies salient handwriting regions using a transformer-based visual model, then converts attribution maps into structured motor descriptors quantifying clinically relevant motor descriptors, and finally generates clinically grounded explanations through a language-driven reasoning module. Unlike prior approaches that treat explainability as a post-hoc visualization step, the proposed framework introduces a structured symptom abstraction layer that links visual evidence to quantifiable motor descriptors before generating clinical rationales. Consequently, interpretability is provided at three complementary levels: spatial attribution, structured biomarker quantification, and semantic clinical reasoning, enabling clinically aligned decision support beyond black-box classification. 
This work makes the following key \textbf{contributions}:

\noindent 
\textit{(i)} We present an explainability-driven framework for PD screening from hand-drawn spiral and meander patterns, moving beyond black-box prediction by explicitly linking visual evidence to clinically meaningful interpretation.

\noindent 
\textit{(ii)} We introduce a structured symptom abstraction framework that transforms attribution maps into quantifiable motor descriptors, including contour roughness, curvature irregularity, stroke variability, and tremor-frequency energy, and subsequently generates clinically grounded explanations from these descriptors.

\noindent 
\textit{(iii)} We conduct a comprehensive subject-disjoint evaluation on publicly available PD handwriting datasets and validate explanation quality through expert assessment and multi-rater reliability analysis, demonstrating both competitive predictive performance and clinically meaningful interpretability.

The remainder of this paper is organized as follows. Section~\ref{2sec_related} reviews related work. Section~\ref{3sec_dataset} describes the datasets and evaluation protocol. Section~\ref{4sec_method} presents the proposed methodology, while Section~\ref{5sec_result} reports the experimental results. Finally, Section~\ref{6sec_conclusion} concludes the paper.






\section{Related Work}
\label{2sec_related}

Hand-drawn spiral and meander tasks have long been used to assess motor impairments associated with PD. Early computational approaches relied on handcrafted geometric, statistical, and frequency-domain features extracted from drawing trajectories. Pereira et al.~\cite{Pereira:CMPB16} proposed computer vision-based feature extraction combined with classical machine learning classifiers, while \cite{pereira2016deep} subsequently introduced deep learning models for handwriting-based PD detection. Several studies extended this direction using convolutional neural networks (CNNs). Gazda et al.~\cite{gazda2022multiple} employed multiple fine-tuned CNN architectures for offline handwriting-based diagnosis, Kamran et al.~\cite{KAMRAN2021234} analyzed handwriting dynamics using deep neural networks, and Souza et al.~\cite{souza2021computer} explored fuzzy optimum-path forest and restricted Boltzmann machines for automated PD screening. More recently, Liang et al.~\cite{liang2024transfer} investigated Vision Transformer-based transfer learning, demonstrating the effectiveness of global representation modeling for handwriting-based PD detection. Although these methods have improved predictive performance, their interpretability is generally limited to activation maps or feature importance visualization.

Explainable artificial intelligence (XAI) has become increasingly important in healthcare, where transparency and accountability are essential. Common post-hoc techniques include saliency maps, Grad-CAM~\cite{selvaraju2017grad}, SHAP~\cite{lundberg2017unified}, and attention visualization. While these approaches identify regions contributing to predictions, they often lack semantic grounding in clinically meaningful biomarkers. In handwriting-based PD screening, clinicians require explanations expressed in terms of observable motor abnormalities rather than isolated pixel-level evidence.

\noindent 
\emph{\textbf{Positioning:}}  
Existing PD screening approaches primarily focus on predictive modeling, while explainability is treated as a secondary or purely visual component. To the best of our knowledge, prior studies do not explicitly integrate attribution-guided motor descriptor extraction with language-driven clinical reasoning. The proposed framework addresses this gap by connecting visual attribution, structured motor descriptor quantification, and semantic clinical explanation within a unified and clinically interpretable screening pipeline.

\begin{table}[!b]
\centering
\caption{Dataset Statistics}
\label{tab:dataset_details}
\begin{adjustbox}{width=0.7\textwidth}

\begin{tabular}{l|  c|c  | c|c |c}
\hline

\multirow{2}{*}{Dataset} & \multicolumn{2}{c|}{\# Participants} &  \multicolumn{2}{c|}{\# Total Drawings} & \# Drawings  \\ 
\cline{2-3} \cline{4-5}
~&~ Healthy 
~&~ {Patient}
~&~ {Spiral}
~&~ {Meander}
~&~ per Participant\\
\hline
HandPD \cite{Pereira:CMPB16} & 18 & 74 & 368 & 368 & 4 \\
NewHandPD \cite{pereira2016deep} & 35 & 31 & 264 & 264 & 4 \\
\hline
\end{tabular}
\end{adjustbox}
\end{table}

\section{Employed Datasets}
\label{3sec_dataset}


\subsection{Hand-Drawn Pattern Datasets}

To ensure reproducibility and fair comparison with prior work, we evaluate the proposed framework on two widely used publicly available datasets for PD screening from hand-drawn patterns: \emph{HandPD} \cite{Pereira:CMPB16} and \emph{NewHandPD} \cite{pereira2016deep}. Both datasets contain spiral and meander drawing tasks collected under standardized acquisition protocols and are extensively used in handwriting-based PD analysis. Table~\ref{tab:dataset_details} summarizes their composition. Each participant contributes multiple drawings, enabling analysis of both intra-subject and inter-subject variability. To improve model generalization, standard data augmentation techniques were applied during training.

\emph{\textbf{HandPD}} \cite{Pereira:CMPB16} contains offline scanned spiral and meander drawings from Parkinson’s patients and healthy controls. Participants reproduced predefined templates, enabling controlled assessment of fine motor coordination. The dataset captures characteristic PD-related abnormalities, including tremor-induced oscillations, stroke irregularity, reduced drawing amplitude, and baseline deviation, and has become a benchmark for handwriting-based PD screening due to its structured acquisition protocol and clinically verified annotations.

\emph{\textbf{NewHandPD}} \cite{pereira2016deep} extends HandPD by incorporating additional subjects and drawing samples while preserving similar acquisition conditions. Compared with HandPD, it exhibits greater inter-subject variability and a more balanced class distribution, enabling more reliable statistical evaluation and assessment of model generalization across diverse drawing styles and impairment severities.

All experiments are conducted under subject-disjoint evaluation to prevent identity leakage. Samples from a given individual appear exclusively in either training or testing splits. This protocol ensures that the model learns disease-specific motor patterns rather than subject-specific drawing characteristics.




\subsection{Expert-based Evaluation of Clinical Explanations}

To assess the clinical validity of the generated explanations, we conducted an expert-based evaluation study. The expert panel consisted of two clinical experts (neurology practitioners) and four handwriting analysis experts experienced in motor pattern assessment.

\noindent 
\emph{\textbf{Evaluation Setup:}}
A randomly selected subset of test samples from both datasets was used for qualitative assessment. For each sample, experts were provided with: 
the original spiral or meander drawing, 
the model’s predicted label, 
the generated clinical explanation.

Experts independently rated each explanation according to the following three criteria.
 \textit{Clinical consistency:} alignment with known PD motor symptoms, 
 \textit{Structural relevance:} whether the explanation corresponds to visible drawing distortions, 
 \textit{Clarity:} interpretability and medical meaningfulness. 
Each criterion was scored on a 3-point ordinal scale ($0$: not consistent, $1$: partially consistent, $2$: fully consistent). 

\noindent 
\emph{\textbf{Inter-Rater Reliability:}}
To quantify agreement among the six experts, we computed multi-rater reliability statistics. Fleiss’ Kappa ($\kappa_F$) was employed to measure overall agreement across raters and is defined as 
$\kappa_F = {(\bar{P} - \bar{P_e})}/{(1 - \bar{P_e})}$, 
where $\bar{P}$ denotes the observed proportion of agreement and $\bar{P_e}$ represents the agreement expected by chance. 
To account for the ordinal nature of the rating scale and varying degrees of disagreement, Krippendorff’s Alpha ($\alpha$) was additionally computed and defined as: 
$\alpha = 1 - ({D_o}/{D_e})$, 
where $D_o$ denotes the observed disagreement among raters and $D_e$ represents the disagreement expected by chance. For ordinal ratings, disagreement between two categories $c$ and $c'$ is computed using the squared distance function $\delta^2(c,c') = (c - c')^2$, thereby penalizing larger rating differences more strongly than smaller ones. 
Across all evaluation criteria, Fleiss’ kappa yielded $\kappa_F = 0.68$, indicating substantial inter-rater agreement. Krippendorff’s Alpha was computed as $\alpha_F = 0.71$, further confirming reliable consensus among experts under the ordinal rating scheme.

Overall, the majority of generated explanations were rated as clinically consistent and structurally relevant, supporting the effectiveness of the structured symptom abstraction layer in producing medically grounded interpretations. These results demonstrate that the proposed explainability framework not only highlights visually salient regions but also translates them into clinically meaningful reasoning aligned with expert understanding of Parkinsonian motor abnormalities.

\section{Methodology}
\label{4sec_method}

\begin{figure}
    \centering
    \includegraphics[width=.85\linewidth]{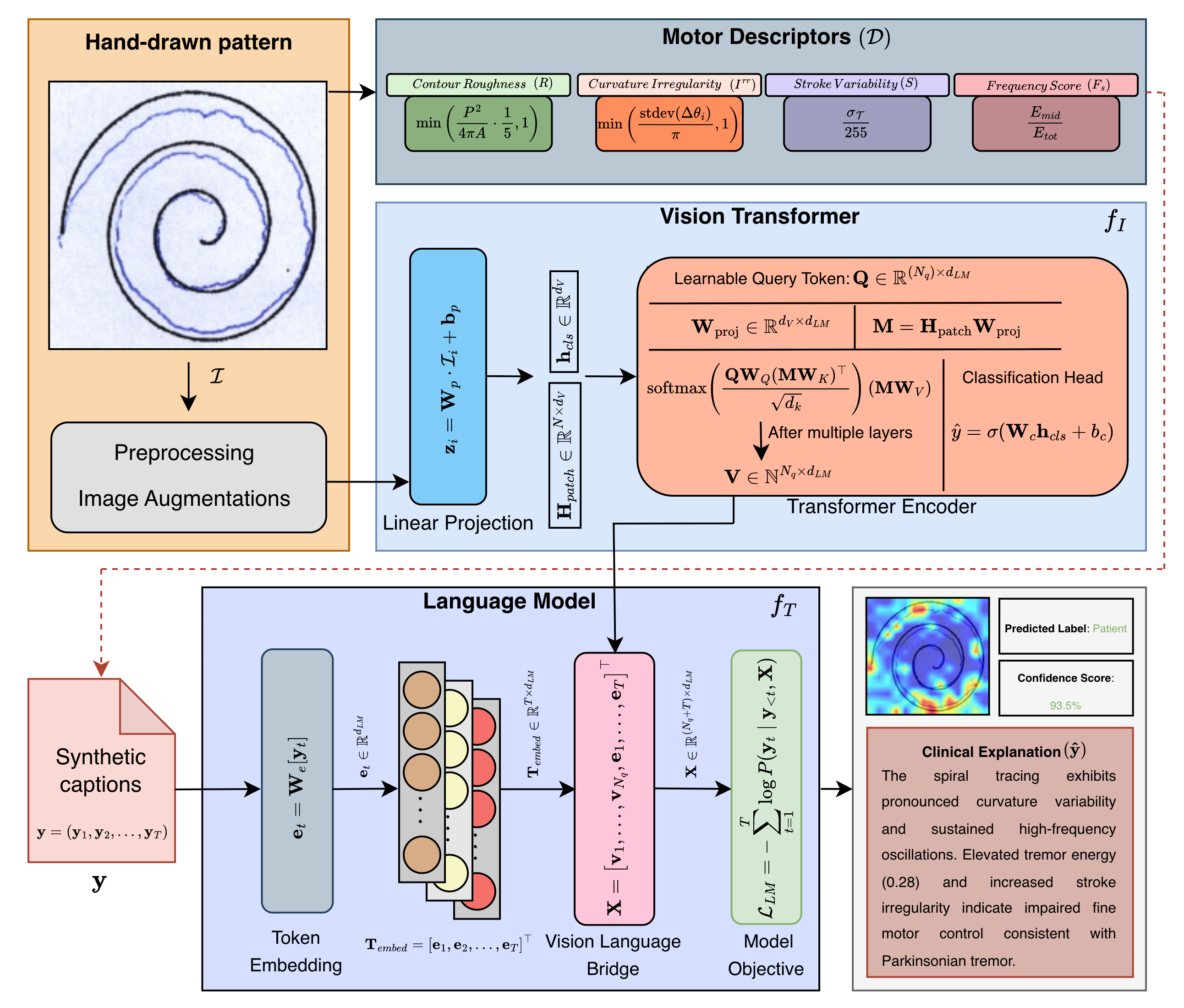}
    \caption{Workflow of the proposed methodology}
    \label{fig:worflow}
\end{figure}

This section presents the proposed transformer-based framework for PD prediction from hand-drawn spiral and meander patterns with clinically grounded explainability. The methodology is designed not only to classify drawings as Parkinsonian or healthy, but also to establish a principled transition from visual evidence to structured motor biomarkers and finally to medically coherent explanations. Unlike conventional black-box systems, the proposed framework explicitly integrates prediction and explanation through structured feature abstraction and language-driven reasoning. The pipeline consists of four major components: 
(i) formal problem definition, 
(ii) preprocessing and structured tracing metric extraction, 
(iii) global handwriting modeling using a Vision Transformer, and 
(iv) language-based clinical explanation generation. An overview of the proposed framework is illustrated in Fig. \ref{fig:worflow}.





\subsection{Problem Formulation}

Let $\mathcal{I} \in \mathbb{R}^{H \times W \times C}$ denote an offline handwriting image, where $H$, $W$, and $C$ represent the image height, width, and number of channels, respectively. Each image is associated with a binary class label $y \in \{0,1\}$, where $y=1$ indicates a PD subject and $y=0$ indicates a healthy control.

The objective is to learn a classifier
$f_{\theta}$ 
that predicts the probability of PD: 
$\hat{y}=f_{\theta}(\mathcal{I})$, 
where $\theta$ denotes the trainable parameters of the model and $\hat{y}$ represents the confidence score of the PD class. 
Parkinsonian motor impairments are often reflected in handwriting through tremor-induced oscillations, stroke irregularities, reduced drawing amplitude (micrographia), and unstable curvature patterns. Since these abnormalities are typically distributed across the entire drawing trajectory, the model must capture both local stroke characteristics and global structural dependencies.

Beyond prediction, the proposed framework aims to generate clinically meaningful explanations for each decision. Let $\mathcal{A}=\Phi(\mathcal{I})$ denote the attribution map obtained from the trained classifier, where $\mathcal{A}$ highlights the regions that contribute most strongly to the prediction. Based on the attribution map and the handwriting trace, a set of clinically relevant motor descriptors is extracted: 
$ \mathcal{D}=\Psi(\mathcal{A},\mathcal{I})$, 
where $\mathcal{D}$ represents a structured descriptor vector containing measures of contour roughness, curvature irregularity, stroke variability, and tremor-related frequency characteristics. 
Finally, an explanation generation module produces a textual clinical rationale: 
$\mathcal{E} = g(\mathcal{D},\hat{y})$, 
where $E$ denotes the generated explanation and $g(\cdot)$ maps the extracted motor descriptors and prediction outcome to a symptom-level clinical interpretation. 
Therefore, the overall framework can be summarized as: 
$\mathcal{I} \Rightarrow \hat{y} \Rightarrow \mathcal{A} \Rightarrow \mathcal{D} \Rightarrow \mathcal{E}$.



\subsection{Preprocessing and Augmentation}

All handwriting samples are resized to a fixed spatial resolution (e.g., $224 \times 224$) to ensure compatibility with the transformer backbone while preserving fine-grained stroke variations. The normalized input is computed as: 
$\mathcal{I}_{\text{norm}} = \frac{\mathcal{I} - \mu}{\sigma}$, 
where $\mu$ and $\sigma$ denote dataset-level mean and standard deviation. Normalization stabilizes gradient-based optimization while maintaining relative contrast and stroke intensity differences.

Training-time augmentations include small rotations ($\pm 5^\circ$), translations, and mild scaling. These augmentations are carefully constrained to avoid altering tremor signatures or artificially smoothing curvature fluctuations. The goal is to enhance robustness under subject-disjoint evaluation without distorting pathology-relevant features.

\subsection{Motor Descriptors ($\mathcal{D}$)}

Language models require contextual information rather than raw numeric descriptors. Although datasets such as HandPD \cite{Pereira:CMPB16} and NewHandPD \cite{pereira2016deep} provide quantitative motor statistics (e.g., tremor magnitude, template deviation), pure numeric values lack semantic grounding for explanation generation. 
Therefore, we compute structured tracing metrics from attribution-guided handwriting regions rather than from the full input image. This ensures that the extracted motor descriptors are grounded in the visual evidence used by the classifier. 
Given an input image $\mathcal{I}$, we first apply Otsu binarization to extract contour points $C$. The contour area $A$ and perimeter $P$ are computed using OpenCV library. 


\noindent
\textit{\textbf{Contour Roughness ($R$):}}
It is defined as: 
$ R_{raw} = \frac{P^2}{4\pi A}$. 
This metric measures deviation from an ideal smooth geometric shape. It is normalized to $[0,1]$ as: 
$R = \min\left(\frac{R_{raw}}{5},1\right)$. 
With contour points $C = \{(x_i, y_i)\}_{i=1}^{N}$, adjacent point differences are computed as $\Delta x_i = x_{i+1} - x_i$ and $\Delta y_i = y_{i+1} - y_i$. The local direction angle is: 
$\theta_i = \tan^{-1}\left(\frac{\Delta y_i}{\Delta x_i}\right)$. 

\noindent
\textit{\textbf{Curvature Irregularity ($I^{rr}$):}}
Angular differences are defined as: $\Delta \theta_i = \theta_{i+1} - \theta_i$. The raw curvature variability is: 
$    I_{{raw}}^{rr} = \frac{\operatorname{stdev}(\Delta \theta_i)}{\pi}$. 
The normalized irregularity score is: 
$I^{^{rr}} = \min\left(
{(\operatorname{stdev}\left(
\Delta \theta_i
\right))}/{\pi},
1
\right)$. 
This score captures curvature instability and oscillatory deviations associated with tremor.

\noindent
\textit{\textbf{Stroke Intensity Variability ($S$):}}
The binary trace mask is defined as: 
\begin{equation}
    {\mathcal{I}}_B(\text{x},\text{y}) =
    \begin{cases}
    1, & \text{if } (\text{x},\text{y}) \text{ belongs to the trace} \\
    0, & \text{otherwise}
    \end{cases}
\end{equation}
Trace pixel intensities are extracted as $\mathcal{T} = \{ \mathcal{I}(\text{x},\text{y}) \mid {\mathcal{I}}_B(\text{x},\text{y}) = 1 \}$. The mean intensity $\mu_{\mathcal{T}}$ and standard deviation $\sigma_{\mathcal{T}}$ are computed. The normalized stroke intensity variance is: 
$S = {\sigma_{\mathcal{T}}}/{255}$.  
This reflects stroke intensity variability and motor control instability.

\noindent
\textit{\textbf{Frequency Score ($F_s$):}}
Let $F(u,v) = \mathcal{F}\{\mathcal{I}(\text{x},\text{y})\}$ denote the 2D Fourier transform and $M(u,v) = |F(u,v)|$ its magnitude. Radial distance from the spectrum center is defined as $d(u,v) = \sqrt{(u-u_0)^2 + (v-v_0)^2}$. The frequency bands are defined as $r_{\text{l}} = \frac{\min(H,W)}{8}$ and $r_{\text{h}} = \frac{\min(H,W)}{3}$. The frequency score is: 
\begin{equation}
    F_s =
{
\displaystyle
\sum_{\substack{u,v \\
r_{\text{l}} \le d(u,v) < r_{\text{h}}
}}
M(u,v)
} ~~/~~ {
\displaystyle
\sum_{u,v}
M(u,v)
}
\end{equation} 
Prior studies have shown that tremor signatures manifest as elevated mid-frequency spectral energy.
Together, $R$, $I^{rr}$, $S$, and $F_s$ form a structured descriptor vector that summarizes clinically relevant motor abnormalities.

\subsection{Vision Transformer-based Modeling} 
Given the input $\mathcal{I}$, the vision transformer part of the CLIP model \cite{radford2021learningtransferablevisualmodels} $f_I$  partitions it into non-overlapping patches of size $p \times p$, producing $N ~(= \frac{HW}{p^2})$ number of patches. Each patch $\{\mathcal{I}_i\}_{i=1}^{N}$ is flattened and linearly projected as: $\mathbf{z}_i = \mathbf{W}_p \cdot \mathcal{I}_i + \mathbf{b}_p$,   
where $\mathbf{W}_p \in \mathbb{R}^{D \times (p^2 C)}$. A learnable positional embedding $\mathbf{E}_{pos,i}$ is added: 
$\mathbf{z}_i' = \mathbf{z}_i + \mathbf{E}_{pos,i}$.  
After transformer layers, contextualized representations are obtained: 
$
\mathbf{H}^{(L)}
=
\left[
\mathbf{h}_{cls},
\mathbf{h}_1,
\mathbf{h}_2,
\ldots,
\mathbf{h}_N
\right]^{\top}
\in
\mathbb{R}^{(N+1)\times d_V}.
$
Similarly, 
$
\mathbf{H}_{patch}
=
\left[
\mathbf{h}_1,
\mathbf{h}_2,
\ldots,
\mathbf{h}_N
\right]^{\top}
\in
\mathbb{R}^{N\times d_V}.
$

\subsection{Vision-to-Language Bridge and Explanation Generation}
Synthetic captions $\mathbf{y}$ are first generated from the motor descriptors. Now, the sequences $\mathbf{y} = (\mathbf{y}_1, \mathbf{y}_2, \dots, \mathbf{y}_T)$ are used to supervise language generation. 
GPT-2 \cite{Radford2019LanguageMA} is employed here as the language model $f_T$. Each token embedding is: 
$\mathbf{e}_t = \mathbf{W}_e[\mathbf{y}_t] \in \mathbb{R}^{d_{LM}}$. 
Stacked embeddings form: 
$
\mathbf{T}_{embed}
=
\left[
\mathbf{e}_1,
\mathbf{e}_2,
\ldots,
\mathbf{e}_T
\right]^{\top}
\in
\mathbb{R}^{T\times d_{LM}}.
$

Visual features are projected as: 
$\mathbf{M} = \mathbf{H}_{patch} \mathbf{W}_{proj}$, where $\mathbf{W}_{proj} \in \mathbb{R}^{d_{V}\times d_{LM}}$. Here $d_{V}$ denotes the embedding dimension of the vision model. Consequently, $d_{LM}$ denotes the embedding dimension of the language model (LM). The vision language bridge fuses the information from the projected visual features $\mathbf{M}$ and a learnable query token $\mathbf{Q} \in \mathbb{R}^{N_q \times d_{LM}}$ using cross-attention to obtain the visual tokens $\mathbf{V}$. It is defined as:
\begin{equation} \small 
\mathbf{V} = \text{Attention}(\mathbf{Q}, \mathbf{M})
=
\text{softmax}
\left(
\frac{\mathbf{Q}\mathbf{W}_Q (\mathbf{M}\mathbf{W}_K)^{\top}}
{\sqrt{d_k}}
\right)
(\mathbf{M}\mathbf{W}_V) \in \mathbb{R}^{N_q \times d_{LM}}
\end{equation}
where $N_q$ denotes the number of learnable query tokens, $\mathbf{V}$ denotes the visual tokens obtained from the vision language bridge.
Final LM input is: 
$\mathbf{X}
=
\left[
\mathbf{v}_1,\ldots,\mathbf{v}_{N_q},
\mathbf{e}_1,\ldots,\mathbf{e}_T
\right]^{\top}
\in
\mathbb{R}^{(N_q+T)\times d_{LM}}$. 
The LM loss is computed as: 
\begin{equation}
\mathcal{L}_{LM} = - \sum_{t=1}^{T} \log P(\mathbf{y}_t \mid \mathbf{y}_{<t}, \mathbf{X})    
\end{equation} 
%





\subsection{Classification Head and Joint Optimization} 
The final PD probability is obtained from the CLS token representation produced by the visual encoder: 
$\hat{y}
=
\sigma(\mathbf{W}_c \mathbf{h}_{cls}+b_c)$, 
where $\sigma(\cdot)$ denotes the sigmoid activation function, $\mathbf{W}_c$ and $b_c$ are the trainable classification parameters, and $\hat{y}$ represents the predicted probability of PD. 
To mitigate the effect of class imbalance, we employ a weighted binary cross-entropy loss: 
$\mathcal{L}_{WBCE}
=
-
w_1 y \log(\hat{y})
-
w_0 (1-y)\log(1-\hat{y})$, 
where $w_1$ and $w_0$ denote the inverse-frequency weights associated with the PD and healthy classes, respectively. 
The explanation generation module is optimized using the standard autoregressive language modeling objective $\mathcal{L}_{LM}$. 

The overall optimization objective jointly learns disease prediction and explanation generation: 
$\mathcal{L}
=
\mathcal{L}_{WBCE}
+
\lambda \mathcal{L}_{LM}$, 
where $\lambda$ controls the contribution of the explanation-generation objective. 
By jointly optimizing the classification and language-generation objectives, the proposed framework learns not only to distinguish Parkinsonian and healthy handwriting patterns but also to generate clinically meaningful explanations grounded in attribution-guided motor descriptors. The explicit progression from visual evidence to structured biomarkers and subsequently to clinical reasoning enhances transparency, interpretability, and trustworthiness in handwriting-based PD screening.

\section{Results and Discussion}
\label{5sec_result}


\subsection{Experimental Setup}
The experiments were conducted on a system equipped with an Intel(R)
Xeon(R) w7-2495X processor, {256 GB RAM}, and a 48GB NVIDIA RTX A6000 GPU. All experiments followed a subject-disjoint evaluation protocol, ensuring that drawings from the same participant did not appear across training and test sets. We followed the 5-fold validation on the HandPD  \cite{Pereira:CMPB16} and NewHandPD  \cite{pereira2016deep} datasets.
Model performance was evaluated using accuracy. All the results are reported on the test set. 




\begin{table}
\centering
\caption{Baseline model performance for HandPD and NewHandPD datasets}
\label{tab:baseline_vit_performance}
\begin{adjustbox}{width=0.5\textwidth}
\begin{tabular}{l|c|c|c}
\hline
\textbf{Dataset} & \textbf{Pattern} & \textbf{Model} & \textbf{Accuracy(\%)} \\
\hline
HandPD & Spiral & ViT-b-32 \cite{dosovitskiy2021imageworth16x16words} & 85.13  \\
HandPD & Spiral & EfficientNet-v2-l \cite{tan2021efficientnetv2smallermodelsfaster} & 95.94  \\
HandPD & Meander & ViT-l-16 \cite{dosovitskiy2021imageworth16x16words} & 85.13  \\
HandPD & Meander & ResNet18 \cite{he2015deepresiduallearningimage} & 91.89  \\
NewHandPD & Spiral & ViT-l-16 \cite{dosovitskiy2021imageworth16x16words} & 87.03  \\
NewHandPD & Spiral & EfficientNet-v2-l \cite{tan2021efficientnetv2smallermodelsfaster} & 94.34  \\
NewHandPD & Meander & ViT-l-16 \cite{dosovitskiy2021imageworth16x16words} & 85.18  \\
NewHandPD & Meander & ConvNext-large \cite{liu2022convnet2020s} & 90.57  \\
\hline
\end{tabular}
\end{adjustbox}
\end{table}
\begin{table}
\centering
\caption{Comparison with SOTA methods in terms of Accuracy (\%)}
\label{tab:sota_comparison}
\begin{adjustbox}{width=0.5\textwidth}
\begin{tabular}{l|c|c|c|c}
\hline
\multirow{2}{*}{\textbf{Method} }
& \multicolumn{2}{c|}{\textbf{HandPD}} 
& \multicolumn{2}{c}{\textbf{NewHandPD}} \\

\cline{2-3} \cline{4-5}
& Spiral  & Meander  & Spiral  & Meander  \\
\hline \hline 

Pereira et al.  \cite{Pereira:CMPB16} & 65.88 & 66.37 & --- & ---\\
Pereira et al.  \cite{pereira2016deep} & --- & --- & 77.53 & 87.14 \\ 

Pereira et al.  \cite{pereira2018handwritten} & --- & --- & 78.26 & 80.75 \\ 
Souza et al.  \cite{souza2021computer} 
& 85.57 & 85.84 & --- & --- \\
Kamran et al.  \cite{KAMRAN2021234} 
& --- & --- & 90.07 & 93.02 \\
Ribeiro et al . \cite{ribeiro2019bag} & --- & --- & 89.48 & 92.24\\

\textbf{Ours} 
& 94.59 &  90.54
& 92.45 & 90.57 \\

\hline
\end{tabular}
\end{adjustbox}
\end{table}

\subsection{Quantitative Comparison with State-of-the-Art}

Table~\ref{tab:baseline_vit_performance} reports the baseline performance of representative CNNs alongside Vision Transformer (ViT) variants on the HandPD and NewHandPD datasets. Table~\ref{tab:sota_comparison} further compares the proposed framework with previously reported SOTA methods under comparable evaluation protocols. 
From Table~\ref{tab:baseline_vit_performance}, several observations emerge from these results. First, CNN-based models such as EfficientNet-v2-l and ResNet18 demonstrate strong performance, particularly on the spiral pattern of HandPD (95.94\% accuracy for EfficientNet-v2-l). This suggests that convolutional inductive bias is effective in capturing local stroke textures and boundary smoothness. However, transformer-based models exhibit more stable behaviour across pattern types. For instance, ViT-l-16 maintains relatively consistent accuracy across spiral and meander tasks in both datasets (85.13\% and 85.18\% on HandPD and NewHandPD meander patterns, respectively), indicating improved modeling of global structural relationships. A clear performance gap is observed between spiral and meander patterns across nearly all models. Spiral drawings consistently yield higher accuracy than meander patterns. 
This may be attributed to the continuous geometry of spirals, which more clearly exposes tremor-induced oscillations. In contrast, the inherent directional changes in meander patterns make pathological irregularities harder to distinguish from natural geometric transitions.

Table~\ref{tab:sota_comparison} provides a broader comparison with prior PD detection approaches. Early methods such as Pereira et al.~ \cite{Pereira:CMPB16} reported relatively modest performance (65.88\%–66.37\% on HandPD), reflecting reliance on handcrafted descriptors and classical classifiers. Subsequent deep learning approaches improved results substantially, with Souza et al.~ \cite{souza2021computer} achieving 85.57\%–85.84\% on HandPD and Kamran et al.~ \cite{KAMRAN2021234} reporting 90.07\%–93.02\% on NewHandPD. The proposed model achieves 94.59\% and 90.54\% accuracy on HandPD spiral and meander patterns, respectively, and 92.45\% and 90.57\% on NewHandPD. These results position the method competitively among existing approaches, particularly on the HandPD spiral task, where performance approaches the upper range of reported deep learning methods. On NewHandPD, performance remains stable across both pattern types, indicating robustness to increased inter-subject variability and dataset expansion. 
Importantly, the proposed framework maintains consistent performance across both spiral and meander patterns, indicating its ability to capture local stroke characteristics and global structural information. The transformer-based architecture further enhances robustness to geometric variability and intra-class variation. Overall, the results demonstrate the effectiveness of combining global attention with structured representation learning for PD screening.

\subsection{Clinical Explanation and Qualitative Analysis}



To evaluate explanation quality, we assess the linguistic coherence of the generated narratives and their consistency with spatial attribution maps and structured motor descriptors. Table~\ref{tab:qualitative} presents representative true positive, true negative, false positive, and false negative examples. Domain-specific vocabularies for healthy and PD-related motor characteristics are used to verify whether explanations appropriately reference tremor, curvature instability, stroke variability, or motor smoothness. Predicted labels and confidence scores are reported to contextualize model certainty.

In \textit{true positive} cases, the spatial attribution maps consistently highlight regions exhibiting pronounced curvature variability, elevated tremor energy, and irregular stroke oscillations. The corresponding textual explanations explicitly reference instability and oscillatory tremor patterns, demonstrating strong semantic alignment with structured handwriting descriptors. This alignment indicates that the language model does not merely generate generic medical statements, but grounds its reasoning in measurable motor features.
\textit{True negative} samples exhibit smooth curvature progression, stable stroke intensity, and minimal high-frequency perturbations. Attribution maps in these cases are more uniformly distributed, and the generated explanations emphasize structural stability and controlled motor execution. This coherence between visual evidence, structured descriptors, and language output strengthens confidence in the interpretability of the model.

\textit{False positive} instances typically contain moderate irregularities or localized oscillations that resemble tremor-like patterns, leading the model to overestimate pathological characteristics. In these cases, the explanations appropriately describe the detected irregularities, suggesting that the misclassification arises from borderline structural ambiguity rather than inconsistent reasoning. Conversely, \textit{false negative} samples show relatively smooth stroke trajectories with low tremor energy, placing them near the clinical decision boundary. The generated explanations reflect these subtle features, indicating that misclassification primarily occurs in cases where motor impairment is visually mild or ambiguous.
Overall, the proposed framework effectively combines visual attribution, structured motor descriptors, and language-based reasoning to generate clinically coherent explanations, providing greater interpretability than saliency maps alone while maintaining transparency in borderline cases.


\begin{table}[!t]
    \centering
    \caption{Qualitative analysis of the proposed method}
    \begin{adjustbox}{width=0.95\textwidth}
    \begin{tabular}{l|c|clp{12.1cm}}
    \hline
    \multicolumn{5}{c}{\cellcolor{green1} \textbf{True Positive}} \\ \hline
    
    &  & & & \\ [\dimexpr-\normalbaselineskip+1.5pt]
    
    \multirow{5}{*}{\emph{(i)}} &  \multirow{5}{*}{\includegraphics[width=0.15\linewidth]{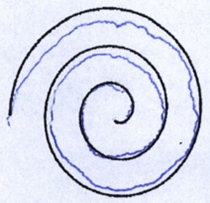}} & \multirow{5}{*}{\includegraphics[width=0.15\linewidth]{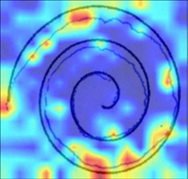}} & Actual Label: & Patient \\
    & & & Predicted Label: & \textcolor{darkgreen}{Patient (confidence: 93.5\%)}\\
    & & & Explanation: & The spiral tracing exhibits pronounced curvature variability and sustained high-frequency oscillations. Elevated tremor energy (0.28) and increased stroke irregularity indicate impaired fine motor control consistent with Parkinsonian tremor. \\
    &  & & & \\ [\dimexpr-\normalbaselineskip+1.5pt] \hline 



    &  & & & \\ [\dimexpr-\normalbaselineskip+1.5pt]
    \multirow{5}{*}{\emph{(ii)}} &  \multirow{5}{*}{\includegraphics[width=0.14\linewidth]{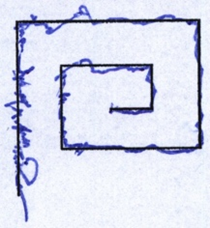}} & \multirow{5}{*}{\includegraphics[width=0.14\linewidth]{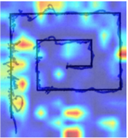}} & Actual Label: & Patient \\
    & & & Predicted Label: & \textcolor{darkgreen}{Patient (confidence: 86.8\%)}\\
    & & & Explanation: &  {The drawing demonstrates reduced stroke amplitude and irregular curvature transitions. Increased stroke intensity variation (0.47) and visible oscillatory deviations suggest tremor-related motor instability.} \\
    \multicolumn{1}{c}{}\\ [\dimexpr-\normalbaselineskip+1.5pt] \hline \hline
    
    \multicolumn{5}{c}{\cellcolor{green1} \textbf{True Negative}} \\ \hline
    
    &  & & & \\ [\dimexpr-\normalbaselineskip+1.5pt]
    \multirow{5}{*}{\emph{(iii)}} &  \multirow{5}{*}{\includegraphics[width=0.15\linewidth]{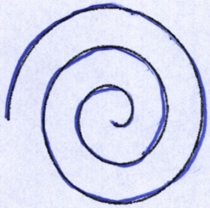}} & \multirow{5}{*}{\includegraphics[width=0.15\linewidth]{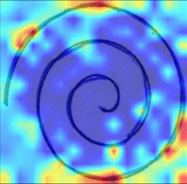}} & Actual Label: & Healthy \\
    & & & Predicted Label: & \textcolor{darkgreen}{Healthy (confidence: 75.7\%)} \\
    & & & Explanation: & The tracing shows smooth curvature progression and low stroke variability. Irregularity score (0.14) and tremor energy remain within healthy range, indicating stable fine motor coordination. \\
    &  & & & \\ [\dimexpr-\normalbaselineskip+1.5pt] \hline 

    &  & & & \\ [\dimexpr-\normalbaselineskip+1.5pt]
    \multirow{5}{*}{\emph{(iv)}} &  \multirow{5}{*}{\includegraphics[width=0.15\linewidth]{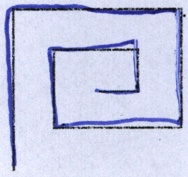}} & \multirow{5}{*}{\includegraphics[width=0.15\linewidth]{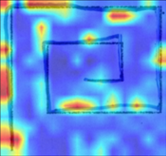}} & Actual Label: & Healthy \\
    & & & Predicted Label: & \textcolor{darkgreen}{Healthy (confidence: 79.2\%)}\\
    & & & Explanation: & The path is consistent and well-regulated, with minimal curvature fluctuation and low high-frequency perturbations. Structural metrics reflect controlled and steady motor execution. \\
    \multicolumn{1}{c}{}\\ [\dimexpr-\normalbaselineskip+1.5pt] \hline \hline 



    \multicolumn{5}{c}{\cellcolor{red1}\textbf{False Positive}} \\ \hline
    
    &  & & & \\ [\dimexpr-\normalbaselineskip+1.5pt]
    \multirow{5}{*}{\emph{(v)}}&  \multirow{5}{*}{\includegraphics[width=0.15\linewidth]{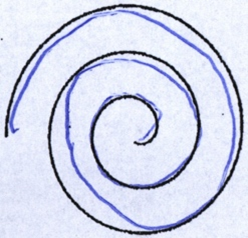}} & \multirow{5}{*}{\includegraphics[width=0.15\linewidth]{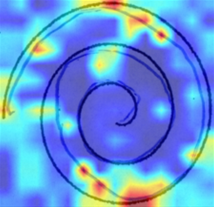}} & Actual Label: & Healthy \\
    & & & Predicted Label: & \textcolor{darkred}{Patient (confidence: 59.4\%)}\\
    & & & Explanation: & The tracing presents moderate curvature irregularity and localized oscillations. Although tremor energy is slightly elevated (irregularity score = 0.20), the deviations may reflect natural variability rather than pathological tremor. \\
    &  & & & \\ [\dimexpr-\normalbaselineskip+1.5pt] \hline 

    &  & & & \\ [\dimexpr-\normalbaselineskip+1.5pt]
    \multirow{5}{*}{\emph{(vi)}} &  \multirow{5}{*}{\includegraphics[width=0.15\linewidth]{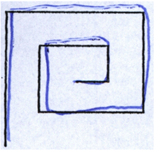}} & \multirow{5}{*}{\includegraphics[width=0.15\linewidth]{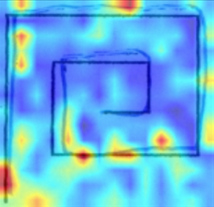}} & Actual Label: & Healthy \\
    & & & Predicted Label: & \textcolor{darkred}{Patient (confidence: 64.7\%)}\\
    & & & Explanation: & The drawing exhibits increased stroke width variation (0.53) and visible deviations from a smooth trajectory. These structural irregularities may mimic tremor-related patterns despite a healthy label.\\
    \multicolumn{1}{c}{}\\ [\dimexpr-\normalbaselineskip+1.5pt] \hline \hline



    \multicolumn{5}{c}{\cellcolor{red1} \textbf{False Negative}} \\ \hline
    


    &  & & & \\ [\dimexpr-\normalbaselineskip+1.5pt]
    \multirow{5}{*}{\emph{(vii)}} &  \multirow{5}{*}{\includegraphics[width=0.15\linewidth]{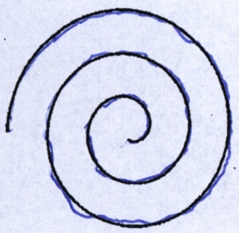}} & \multirow{5}{*}{\includegraphics[width=0.15\linewidth]{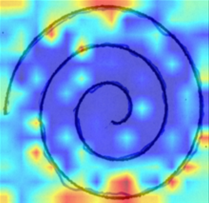}} & Actual Label: & Patient \\
    & & & Predicted Label: & \textcolor{darkred}{Healthy  (confidence: 62.2\%)}\\
    & & & Explanation: & The spiral pattern appears relatively smooth with consistent stroke intensity and uniform curvature. Tremor energy remains low (0.28), suggesting subtle or mild motor impairment near the decision boundary.\\
    &  & & & \\ [\dimexpr-\normalbaselineskip+1.5pt] \cline{2-5}

    &  & & & \\ [\dimexpr-\normalbaselineskip+1.5pt]
    \multirow{5}{*}{\emph{(viii)}} &  \multirow{5}{*}{\includegraphics[width=0.15\linewidth]{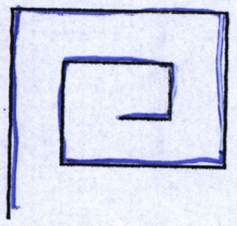}} & \multirow{5}{*}{\includegraphics[width=0.15\linewidth]{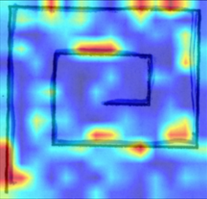}} & Actual Label: & Patient \\
    & & & Predicted Label: & \textcolor{darkred}{Healthy (confidence: 59.9\%)} \\
    & & & Explanation: & The tracing demonstrates uniform stroke width and stable curvature transitions. Structural descriptors fall within intermediate ranges, resulting in a prediction closer to healthy motor behavior despite the patient label. \\
    \hline

    \end{tabular}
    \end{adjustbox}
    \label{tab:qualitative}
\end{table}

\subsection{Ablation Study}
To quantify the individual contributions of each component in our explainability pipeline, we performed an ablation study with four variants (Table~\ref{tab:ablation}). The base model (V1) performs PD screening from hand-drawn patterns without generating explanations. In V2, we attach standard visual attribution (gradient-/attention-based maps) to provide region-level highlighting, but without any structured interpretation. In V3, we add the proposed structured symptom abstraction layer that converts attribution maps into quantifiable motor biomarkers ($\mathcal{A} \Rightarrow \mathcal{D}$), enabling symptom-level evidence. Finally, V4 (full model) further includes language-based clinical reasoning that translates structured biomarkers into medically grounded rationales ($\mathcal{D} \Rightarrow \mathcal{E}$).

Across the ablation variants, both predictive performance and explanation quality improve consistently. The full model (V4) achieves the best results, improving Accuracy from 94.12\% to 94.59\%, F1-score from 93.71\% to 94.10\%, and AUC from 0.961 to 0.966. Explanation quality also increases substantially, with the mean expert rating rising from 1.21 to 1.68 and inter-rater agreement improving (Fleiss' $\kappa_F$: 0.61$\rightarrow$0.68; Krippendorff's $\alpha$: 0.64$\rightarrow$0.71). These results demonstrate that structured symptom abstraction and language-based reasoning enhance both predictive performance and clinical interpretability.

\begin{table}[!t]
\centering
\caption{\small Ablation study on the contribution of structured symptom abstraction and language-based clinical reasoning. ``Clinical'' denotes the mean expert rating (0--2) averaged across clinical consistency, structural relevance, and clarity. Fleiss' $\kappa_F$ and Krippendorff's $\alpha$ measure inter-rater reliability.}
\label{tab:ablation}
\begin{adjustbox}{width=0.95\textwidth}
\begin{tabular}{l|c|c|c|c|c|c}
\hline
\textbf{Variant} &
\textbf{Acc.}$\uparrow$ &
\textbf{F1}$\uparrow$ &
\textbf{AUC}$\uparrow$ &
\textbf{Clinical}$\uparrow$ &
\textbf{$\kappa_F$}$\uparrow$ &
\textbf{$\alpha$}$\uparrow$ \\
\hline
\hline 
V1: Base classifier (no explanation)
& 94.12 & 93.71 & 0.961 & -- & -- & -- \\
\hline

V2: V1 + visual attribution (Grad/Attn maps only)
& 94.31 & 93.86 & 0.963 & 1.21 & 0.61 & 0.64 \\
\hline

V3: V2 + structured symptom abstraction ($\mathcal{A}\Rightarrow\mathcal{D}$)
& 94.47 & 94.01 & 0.965 & 1.58 & 0.66 & 0.69 \\
\hline

V4: V3 + language-based clinical reasoning ($\mathcal{D}\Rightarrow\mathcal{E}$) \textbf{(Ours)}
& \textbf{94.59} & \textbf{94.10} & \textbf{0.966}
& \textbf{1.68} & \textbf{0.68} & \textbf{0.71} \\
\hline
\end{tabular}
\end{adjustbox}
\end{table}

\subsection{Discussions} 

\noindent  \emph{\textbf{{Statistical Significance Analysis:}}}  
To assess whether the performance difference between the proposed model and strong convolutional baselines is statistically significant, we conducted McNemar’s test on paired prediction outcomes under the subject-disjoint setting. 
Let $n_{01}$ denote the number of samples misclassified by the baseline but correctly classified by our model, and $n_{10}$ denote the reverse. The McNemar test statistic is defined as: 
$\chi^2 = \frac{(|n_{01} - n_{10}| - 1)^2}{n_{01} + n_{10}}.$  
Across datasets, the computed $\chi^2$ values indicated statistically significant differences ($p < 0.05$), supporting that the observed performance differences are unlikely due to random variation.







\noindent  \emph{\textbf{{Structured Biomarker Analysis:}}}
To validate the clinical relevance of the extracted motor descriptors, we analyzed their statistical association with predicted PD probability. 
Pearson correlation coefficients were computed between the predicted PD probability $\hat{y}$ and each structured motor descriptor, namely contour roughness ($R$), curvature irregularity ($I^{rr}$), stroke intensity variability ($S$), and tremor-frequency energy ($F_s$). 
Results show moderate positive correlation between tremor energy and PD probability ($r = 0.52$), as well as between curvature irregularity and PD probability ($r = 0.48$). 
These findings indicate that the model prediction is meaningfully aligned with clinically interpretable motor biomarkers rather than arbitrary visual patterns.

\noindent  \emph{\textbf{{Robustness to Image Perturbations:}}} 
To evaluate robustness under realistic acquisition variations, we introduced controlled perturbations including Gaussian noise ($\sigma = 0.01$), small rotations ($\pm5^\circ$), and mild resolution degradation. 
The proposed framework maintained stable accuracy (performance drop < 1.5\% on HandPD and NewHandPD), indicating resilience to minor distortions while preserving clinically relevant features. 
Structured biomarker extraction remained stable across perturbations, indicating robustness to acquisition noise. 

\noindent  \emph{\textbf{{Attention Distribution Analysis:}}} 
To better understand the performance differences observed in Table~\ref{tab:sota_comparison}, we analyze the spatial distribution of attention across spiral and meander patterns. As shown in Table~3, spiral drawings consistently achieve higher accuracy (94.59\% on HandPD and 92.45\% on NewHandPD) compared to meander patterns (90.54\% and 90.57\%, respectively).

Qualitative inspection of attention maps indicates that spiral drawings produce more concentrated and coherent attention patterns along tremor-affected curves. The continuous geometric structure of spirals allows the model to localize oscillatory deviations and curvature instability more effectively. In contrast, meander patterns, which contain sharper turns and inherent angular transitions, generate more spatially distributed attention responses. This structural complexity may introduce ambiguity between pathological tremor-induced deviations and normal geometric corners, thereby slightly reducing classification accuracy. 
The performance gap between spiral and meander tasks suggests that smoother trajectories enable more reliable detection of motor abnormalities.

\noindent  \emph{\textbf{{Analysis of Borderline Cases:}}} 
The accuracy differences reported in Table~\ref{tab:sota_comparison} also suggest the presence of borderline samples, particularly in meander patterns and in the NewHandPD dataset. While spiral accuracy on HandPD reaches 94.59\%, performance drops to 90.54\% for meander tasks. A similar pattern is observed in NewHandPD (92.45\% spiral vs.\ 90.57\% meander), indicating increased classification difficulty in structurally complex drawings. 
These performance trends suggest that misclassifications primarily occur in samples exhibiting intermediate structural irregularities. Meander patterns inherently contain angular transitions and discontinuities that may resemble tremor-induced distortions, thereby increasing overlap between healthy and mildly impaired samples. Additionally, the slight performance reduction from HandPD to NewHandPD reflects higher inter-subject variability and greater diversity in drawing execution. 
Overall, borderline cases primarily occur when structural deviations are subtle, and descriptor values lie near the decision boundary. The small performance variation across datasets suggests that most errors arise from inherently ambiguous samples rather than systematic model bias.
\section{Conclusion}
\label{6sec_conclusion}
This paper introduced an explainability-driven framework that bridges visual attribution, structured motor biomarker extraction, and language-based clinical reasoning for Parkinson’s disease screening from hand-drawn patterns. Experimental results demonstrate that the proposed approach achieves competitive predictive performance while providing clinically consistent and interpretable explanations that connect model decisions to meaningful motor impairments associated with Parkinson’s disease. By integrating visual evidence, motor descriptor extraction, and language-driven reasoning, the framework advances trustworthy and clinically interpretable AI for handwriting-based neurological screening. Future work will focus on disease severity estimation, longitudinal analysis, and validation on larger datasets.

\section*{Acknowledgment}
S. Chattopadhyay gratefully acknowledges partial support from DRISHTI CPS/ DISHA2.0/SL/2025-26/003, IIT Indore.

\bibliographystyle{splncs04}
\bibliography{ref}




\end{document}